# Text Summarization using Deep Learning and Ridge Regression

*Karthik Bangalore Mani*

*Illinois Institute of Technology*

**Abstract**

We develop models and extract relevant features for automatic text summarization and investigate the performance of different models on the DUC 2001 dataset. Two different models were developed, one being a ridge regressor and the other one was a multi-layer perceptron. The hyperparameters were varied and their performance were noted. We segregated the summarization task into 2 main steps, the first being sentence ranking and the second step being sentence selection. In the first step, given a document, we sort the sentences based on their Importance, and in the second step, in order to obtain non-redundant sentences, we weed out the sentences that are have high similarity with the previously selected sentences.

**Introduction**

The process of text summarization is to condense the information as much as possible without losing the gist of the document. In this project, we develop an extractive summarizer which extracts the most important sentences in a document, which are also salient. There are 2 main steps in a summarization task, namely sentence ranking and sentence selection. The first step is done to get an importance score for every sentence in the document and the second step is done to avoid redundancy in the summary, which weeds out sentences that convey the same meaning as the earlier selected sentences.

**Sentence ranking** – We use the predicted ROUGUE-2 scores of the models, and sort them in the descending order. The ones with high ROGUE-2 predictions are considered to be important.

**Sentence selection** – We use a greedy approach (Li and Li 2014) to stitch together multiple sentences for the summary. In each step of selection, the sentence with maximal salience is added into the summary, unless its similarity with a sentence already in the summary exceeds a threshold. Here we use tf-idf cosine similarity and (Cao et. al 2015) set the threshold Tsim = 0.6.

The process of summarization was converted to a regression task wherein the X_Matrix had 9 features for every sentence and Y value was the rouge-2 score between the sentence and the real summary in the DUC dataset. Different models such as deep MLP and ridge were trained and cross validated on this X_Matrix and Y. Their hyperparameters were varied and accuracies were plotted. Due to the limited size of dataset and hand-crafted features, we found that the simple ridge regressor beat all the deep models. Since ridge was the best model, sentences were ranked and selected using ridge regressor.

## 1. Approach
### 1.1 Data Collection
Document Understanding Conference (DUC) is a standard dataset to experiment with and evaluate summarization models. Hence, we collected the DUC 2001 dataset to build the models. This Dataset has 310 documents with complete texts and summaries written by a human.

### 1.2 Feature Extraction
A total of 9 features were extracted for every sentence across every documents. The 9 features are listed below:

1. **Position** - The position of the sentence. Suppose there are M sentences in the document, then for the $i^{th}$ sentence the position is computed as $1-(i-1)/(M-1)$

2. **Length** - No of words in the sentence

3. **Averaged TF** - The mean term frequency of all words in the sentence, divided by the sentence length.

4. **Averaged IDF** - The mean inverse document frequency of all words in the sentence, divided by the sentence length.

5. **Averaged CF** - The mean cluster frequency of all words in the sentence, divided by the sentence length.

6. **POS ratio** – The number of nouns, verbs, adverbs, adjectives in the sentence, divided by the length of the sentence

7. **Named Entity ratio** – The number of named entities in the sentence, divided by the length of the sentence

8. **Number ratio** – The number of digits in the sentence, divided by the length of the sentence

9. **Stopword ratio** – The number of stopwords in the sentence, divided by the length of the sentence. We use the stopword list in the nltk package.

After extracting the above 9 features, the train matrix was constructed with the N*M where,

$$N = \sum_{i=1}^{c} \sum_{j=1}^{d_i} X_{ij}$$

Where c is the no of clusters, $d^i$ is the no of docs in cluster$^i$, $X^{ij}$ is the no of sentences in $j^{th}$ doc of the cluster$^i$ and M = 9, which is the number of features for every sentence.

### 1.3 First Sentence Baseline Model
Usually, it has been argued that the first sentence of the document captures the most important information of the document. Hence a dummy model which blindly predicts the first sentence as the predicted summary was built. The mean ROGUE-2 score between the first sentence and the actual summary across all documents was computed and it's performance was noted.

## 1.4 Ridge Regression Model

Ridge regression (Tibshirani 2013) is like least squares but shrinks the estimated coefficients towards zero. Given a response vector $y \in R^n$ and a predictor matrix $X \in R^{n \times p}$, the ridge regression coefficients are defined as :

$$\hat{\beta}^{ridge} = \underset{\beta \in R^p}{\operatorname{argmin}} \sum_{i=1}^{n}(y_i - x_i^T \beta)^2 + \lambda \sum_{j=1}^{p} \beta_j^2$$

$$= \underset{\beta \in R^p}{\operatorname{argmin}} \underbrace{\|y - X\beta\|_2^2}_{Loss} + \lambda \underbrace{\|\beta\|_2^2}_{Penalty}$$

Here $\lambda \geq 0$ is a tuning parameter, which controls the strength of the penalty term. Note that:
- When $\lambda = 0$, we get the linear regression estimate
- When $\lambda = \infty$, we get $\hat{\beta}^{ridge} = 0$
- For $\lambda$ in between, we are balancing two ideas: fitting a linear model of y on X, and shrinking the coefficients

### 1.4.1 Ridge Validation Error

During the validation phase, we used 10-fold cross validation to identify the best parameters for the regressor. Upon cross validating for various polynomial features such as 1, 2 and 3, we found that the validation error is minimum when the polynomial order is 2, as shown in the below plot –

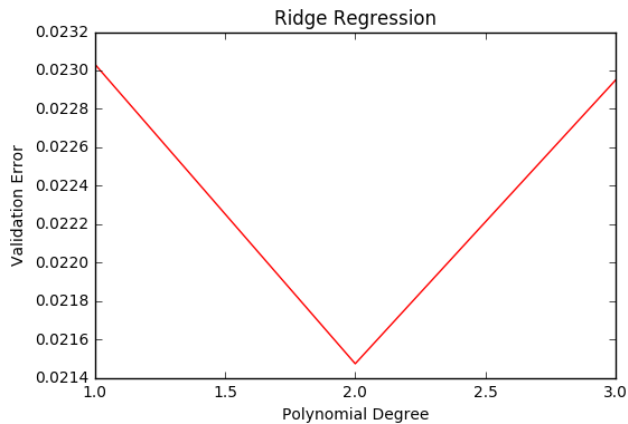

Hence, the polynomial order 2 was chosen and the X_matrix was raised to this order during the testing phase.

## 1.5 Multi-Layer Perceptron

A multilayer perceptron (Wikipedia) is a feedforward artificial neural network model that maps sets of input data onto a set of appropriate outputs. An MLP consists of multiple layers of nodes in a directed graph, with each layer fully connected to the next one. Except for the input nodes, each node is a neuron (or processing element) with a nonlinear activation function. MLP utilizes a supervised learning called backpropagation for training the network.

To put it in simple words, training an MLP has 2 main passes, namely the forward pass and the backward pass. In the forward pass, we compute the output of the activation functions, and in the backward pass, we find the error of the activation functions and finally, we make weight updates.

The weight updates are generally done as follows (Swingler) –

$$\Delta w = w - w_{old} = -\eta \frac{\partial E}{\partial w} = +\eta \delta x$$

So, the weight change from the input layer unit $i$ to hidden layer unit $j$ is:

$$\Delta w_{ij} = \eta \cdot \delta_j \cdot x_i \quad \text{where} \quad \delta_j = o_j(1-o_j)\sum_k w_{jk} \cdot \delta_k$$

The weight change from the hidden layer unit $j$ to the output layer unit $k$ is:

$$\Delta w_{jk} = \eta \cdot \delta_k \cdot o_j \quad \text{where} \quad \delta_k = (y_{target,k} - y_k) y_k (1 - y_k)$$

An MLP will have 1 input layer, 1 output layer and varying hidden layers. For experiments in our project, we had an MLP with the below architecture:

- Input Nodes - A total of 56 input nodes
- Hidden Nodes - A total of 57 hidden nodes
- Output Node – 1 linear output node to get the predicted rouge score

The hidden layers were varied and their performance were noted.

Below is an example of a MLP with 1 input, hidden and output layers.

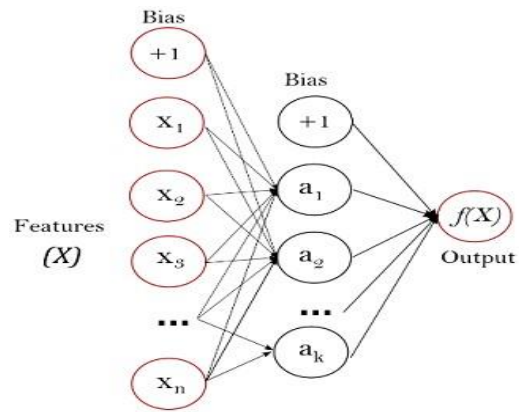

Figure 1 : One hidden layer MLP.

### 1.5.1 MLP Validation Error

We use the same 10-fold cross validation for Deep MLP to tune the hyperparameters. To figure out the best hyperparameters, we validate the model with various settings as show below –

- Epochs – 5, 10, 15, 20
- Optimizers – SGD, Adam, LBFGS
- Activation Functions – Logistic and Tanh
- Hidden Layers – 1, 2, 3

The number of units at every hidden layer was fixed at 57. So, we validated MLP with a total of 4*3*2*3 = 72 different settings. The validation errors for the 72 settings were as shown below –

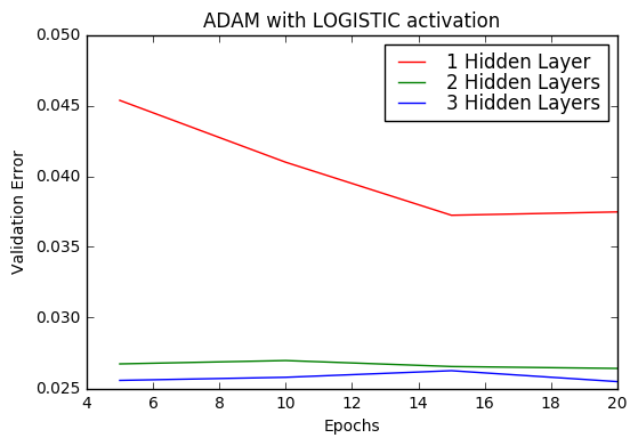
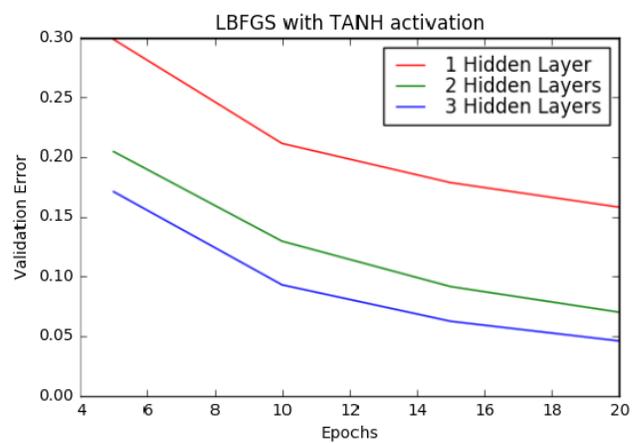
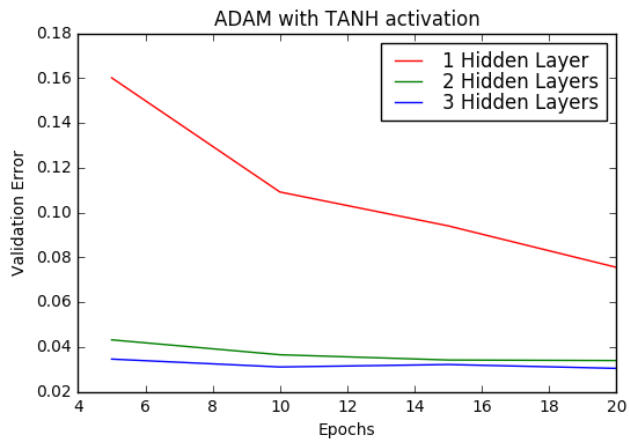
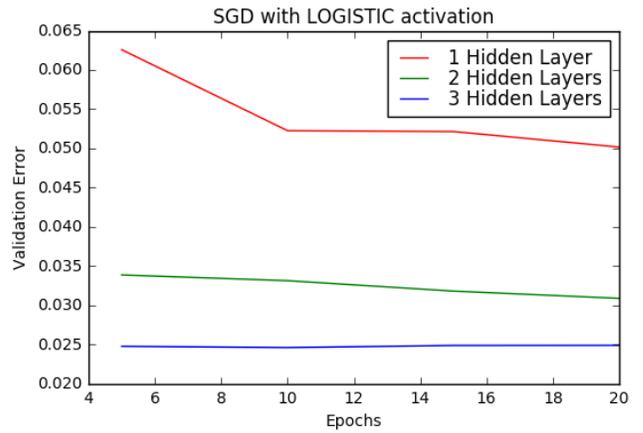
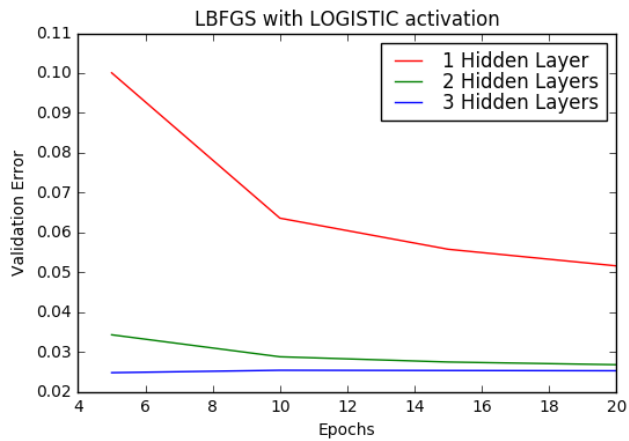
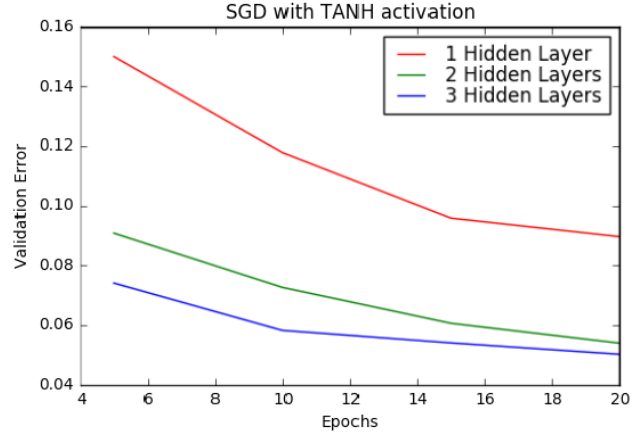

From the above plots, we find the that the validation error is minimum for the 3 optimizers with logistic as the activation function, with 3 hidden layers -

| Optimizer | Activation | Hidden Layers |
|---|---|---|
| Adam | Logistic | 3 |
| LBFGS | Logistic | 3 |
| SGD | Logistic | 3 |

Hence, for all the 3 optimizers, we chose logistic as the activation function with the number of hidden layers to be 3 during the testing phase.

## 2. Results

By using the best settings identified during the validation phase, we fit different models with below settings –

- First sentence model
- Ridge with 2$^{nd}$ order polynomial features
- MLP with 3 hidden layers, with adam optimizer and logistic activation
- MLP with 3 hidden layers, with lbfgs optimizer and logistic activation.
- MLP with 3 hidden layers, with sgd optimizer and logistic activation

Test accuracy is the ROGUE-2 score between the predicted summary and actual gold summary in the DUC dataset. We plot the accuracies as shown below and find that the simple Ridge regression model beats other Deep Models -

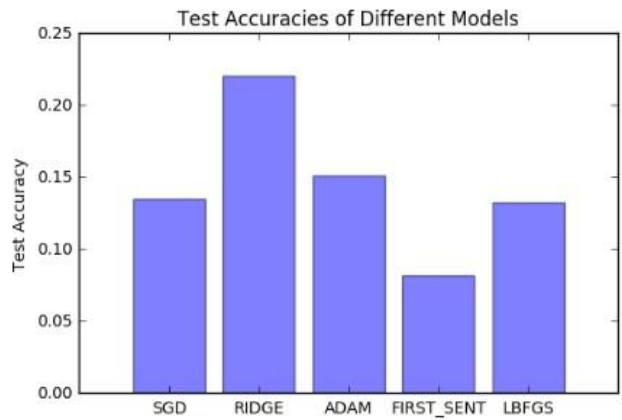

## 3. Future Work

The reason ridge regression beats other models is due to limited size of the dataset we obtained (310 documents), which is actually very low for any deep learning standard. Another reason is that the features were hand-crafted and fed to the models. Hence, in the future we intend to do the following –

- Train on all the DUC datasets 2001,2002….2016
- Learn the features from text instead of handcrafting them
- Fine tune the hyperparameters by using dropout
- Generate an abstract summary instead of extracting it
- Work on query oriented summarization
- Fit other models such as Recurrent Neural Nets.

## 4. REFERENCES

[1] Kevin Swingler. *Multi-Layer Perceptrons*. http://www.cs.stir.ac.uk/courses/ITNP4B/lectures/kms/4-MLP.pdf

[2] *Multilayer Perceptron*. https://en.wikipedia.org/wiki/Multilayer_perceptron

[3] *Neural Network Models (Supervised)*. http://scikit-learn.org/stable/modules/neural_networks_supervised.html

[4] Ryan Tibshirani. *Modern regression*. http://www.stat.cmu.edu/~ryantibs/datamining/lectures/16-modr1.pdf

[5] Ziqiang Cao, Furu Wei, Li Dong, Sujian Li, Ming Zhou. *Ranking with Recursive Neural Networks and Its Application to Multi-Document Summarization*. www.aaai.org/ocs/index.php/AAAI/AAAI15/paper/download/9414/9520